\documentclass[10pt,twocolumn,letterpaper]{article}

\usepackage{iccv}
\usepackage{times}
\usepackage{epsfig}
\usepackage{graphicx}
\usepackage{amsmath}
\usepackage{amssymb}
\usepackage{color}
\usepackage{kotex}
\usepackage{authblk}

\usepackage[pagebackref=true,breaklinks=true,letterpaper=true,colorlinks,bookmarks=false]{hyperref}

 \iccvfinalcopy 

\def\iccvPaperID{9500} 

\ificcvfinal\fi

\begin{document}

\title{LFI-CAM: Learning Feature Importance for Better Visual Explanation}

\author[1,*, **]{Kwang Hee Lee}

\author[1,*]{Chaewon Park}

\author[1,2,*]{Junghyun Oh}

\author[2]{Nojun Kwak}

\affil[1]{Boeing Korea Engineering and Technology Center(BKETC)}
\affil[2]{Seoul National University}



\renewcommand\Authands{ and }

\maketitle
\ificcvfinal\thispagestyle{empty}\fi

\newcommand\blfootnote[1]{%
  \begingroup
  \renewcommand\thefootnote{}\footnote{#1}%
  \addtocounter{footnote}{-1}%
  \endgroup
}

\blfootnote{ \textsuperscript{*} indicates equal contribution.}
\blfootnote{ \textsuperscript{**} indicates corresponding author. }

\begin{abstract}

Class Activation Mapping (CAM) is a powerful technique used to understand the decision making of Convolutional Neural Network (CNN) in computer vision. Recently, there have been attempts not only to generate better visual explanations, but also to improve classification performance using visual explanations. However, the previous works still have their own drawbacks. In this paper, we propose a novel architecture, LFI-CAM, which is trainable for image classification and visual explanation in an end-to-end manner.  LFI-CAM generates an attention map for visual explanation during forward propagation, at the same time, leverages the attention map to improve the classification performance through the attention mechanism. Our Feature Importance Network (FIN) focuses on learning the feature importance instead of directly learning the attention map to obtain a more reliable and consistent attention map. We confirmed that LFI-CAM model is optimized not only by learning the feature importance but also by enhancing the backbone feature representation to focus more on important features of the input image. Experimental results show that LFI-CAM outperforms the baseline models's accuracy on the classification tasks as well as significantly improves on the previous works in terms of attention map quality and stability over different hyper-parameters.

\end{abstract}


\begin{figure}
\begin{center}
\includegraphics[width=1\linewidth, height=0.9\linewidth]{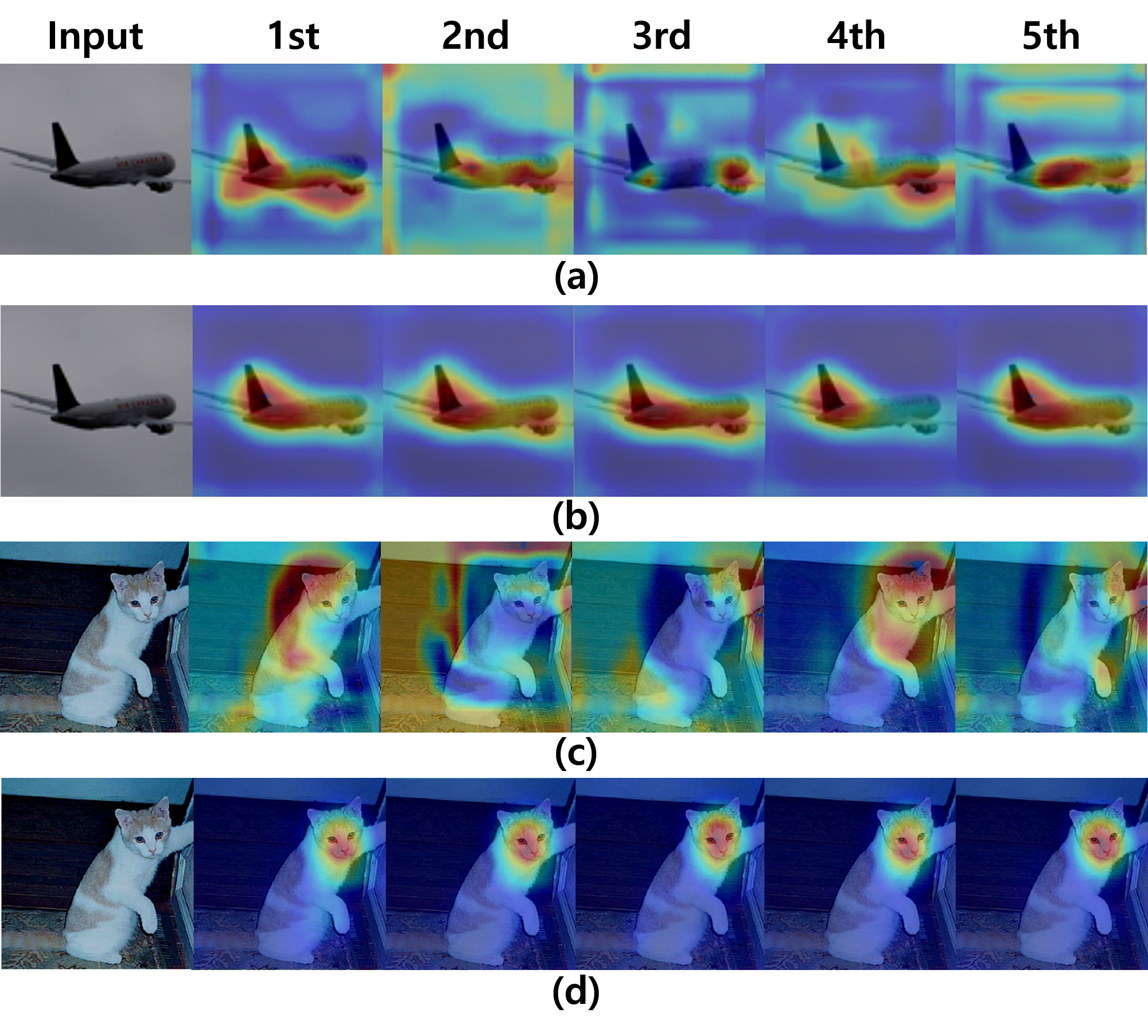}
\end{center}
  \caption{ Examples of stability test on visual explanation. Each row displays CAM results of ABN or LFI-CAM models that were trained with various (5) hyper-parameters. As illustrated, ABN's CAM results are unreliable and inconsistent even for same test images despite the similar accuracies of the models.
  On the other hand, LFI-CAM results in much more consistent CAM images with better visual quality. (a)(c) ABN on STL10 (a) and Cat$\&$Dog (c), (b)(d) LFI-CAM on STL10 (b) and Cat$\&$Dog (d).
 }
\label{fig:Stability_Test}
\end{figure}
\section{Introduction}


As Convolutional Neural Network (CNN) models have become mainstream in computer vision tasks~\cite{antol2015vqa, ren2015faster, he2016deep, long2015fully, vinyals2015show, he2017mask,huang2017densely, krizhevsky2017imagenet}, a rising need to understand the rationale behind the models' decision and prediction has surfaced. Most deep neural networks are considered as black box due to the huge number of parameters and implicit non-linearity. We currently use various metrics such as accuracy, precision, etc. to evaluate the models but sometimes these metrics can be misleading or inaccurate. To empower humans to trust the model, models should be equipped with the capability of providing human-comprehensible explanation on why it made certain decisions. 

To address this need, several visual explanation methods have been proposed~\cite{zeiler2014visualizing, smilkov2017smoothgrad, selvaraju2017grad, zhou2016learning, chattopadhay2018grad, ribeiro2016should, petsiuk2018rise, fukui2019attention, wang2020score, montavon2018methods} and are being widely used for various recognition tasks. These methods include, but are not limited to CAM~\cite{zhou2016learning} , Grad-CAM~\cite{selvaraju2017grad}, Grad-CAM++~\cite{chattopadhay2018grad}, LIME~\cite{ribeiro2016should}, RISE~\cite{petsiuk2018rise}, ABN~\cite{fukui2019attention}, and Score-CAM~\cite{wang2020score}. Broadly speaking, we can categorize the aforementioned methods into 4 categories: response-based, gradient-based, perturbation-based, and hybrid-based visual explanation. 

{\bf Response-based.} CAM~\cite{zhou2016learning} is a response-based visual explanation model which replaces the fully connected layer with Global Average Pooling (GAP) and projects weight matrix onto the channel-wise averaged feature maps. This method however is restrictive as it requires architecture-sensitive alterations in the original network structure, with degradation in classification accuracy compared to non-interpretable models.

{\bf Gradient-based.} Grad-CAM~\cite{selvaraju2017grad}  is a gradient-based visual explanation model that leverages the global average pooling of partial derivatives to capture the importance of a feature map for a target class. It fails to localize multiple occurrences of the same class and the entire region of the object. Grad-CAM++~\cite{chattopadhay2018grad} builds upon Grad-CAM's logic by capturing the weighted average of positive partial gradients to resolve the downsides of Grad-CAM. Both models require an extra back-propagation step during inference time.

{\bf Perturbation-based.} LIME~\cite{ribeiro2016should} applies perturbations on the input to learn a locally-weighted linear regression model that presents image regions as explanation that have the highest positive weight in approximating the true label. Though it is model-agnostic and simple, it requires additional regularization and is time-consuming. RISE~\cite{petsiuk2018rise} estimates the importance of input image regions as the predicted score by randomly sampling masks.

{\bf Hybrid-based.} Score-CAM~\cite{wang2020score} is a hybrid of perturbation-based and response-based model. It uses attention maps as masks on the original image, and a forward-passing score on the target class is obtained and then aggregated as a weighted sum of score-based weights and attention maps. Though it achieves high accuracy and stable results compared to gradient-based methods, Score-CAM is very slow and time-consuming as it needs as many inferences as the number of feature maps to obtain CAM. 
Attention Branch Network (ABN)~\cite{fukui2019attention} is a hybrid model that uses a response-based model with the attention mechanism. It optimizes the loss term, which is the sum of attention loss and perception loss. ABN's limitation is that it often results in an unstable and suboptimal attention map for certain hyper-parameter settings (See Fig.~\ref{fig:Stability_Test}).

Inspired by ABN and Score-CAM, we propose a novel architecture, LFI-CAM, which follows a similar structure to ABN. However, to constrain the attention map generation process as close as possible to the original CAM method, the LFI-CAM attention branch treats the feature maps as masks and obtains feature importance scores for each feature map to generate the attention map in a similar manner to Score-CAM. Unlike Score-CAM, LFI-CAM's Feature Importance Network (FIN) in the attention branch learns the feature importance for each feature map during training. Hence, LFI-CAM's attention map is generated much faster than Score-CAM during forward propagation.

LFI-CAM is composed of two parts: attention branch and perception branch. The attention branch plays an important role in LFI-CAM because it not only generates an attention map for visual explanation by learning the feature importance, but also leverages the attention map to improve the classification performance using the attention mechanism. The perception branch extracts feature maps and predicts a class through the attention mechanism using the feature map from the convolutional layer and attention map. LFI-CAM is trainable for image classification and visual explanation in an end-to-end manner and outputs a more reliable and consistent attention map with smaller network parameters than ABN. Our key contributions in this work are summarized as follows:

(1) We propose a new architecture LFI-CAM for image classification and visual explanation based on Class Activation Mapping with a simple but efficient learnable feature importance for each feature map.

(2) LFI-CAM learns the feature importance of attention maps in an intuitive and understandable way and leverages attention mechanism to improve classification performance and generate more reliable and consistent attention map simultaneously during forward propagation. When compared to Score-CAM, our model is equivalent in visual explanation quality but much faster in speed. Also, it results in better attention map quality and classification accuracy with smaller network parameters compared to ABN.

(3) As a gradient-free method, LFI-CAM bridges the gap between perturbation-based and CAM-based methods with much faster inference speed than Score-CAM.

(4) LFI-CAM is not architecture-sensitive and can be easily applied to various baseline models such as ResNet~\cite{he2016deep}, DenseNet~\cite{huang2017densely}, ResNeXt~\cite{xie2017aggregated} and SENet~\cite{hu2018squeeze} by combining the baseline model with the Feature Importance Network and the attention mechanism.

\section{Preliminary}

\subsection{Attention Branch Network (ABN)}

\begin{figure*}
\begin{center}
\includegraphics[width=1\linewidth, height=0.5\linewidth]{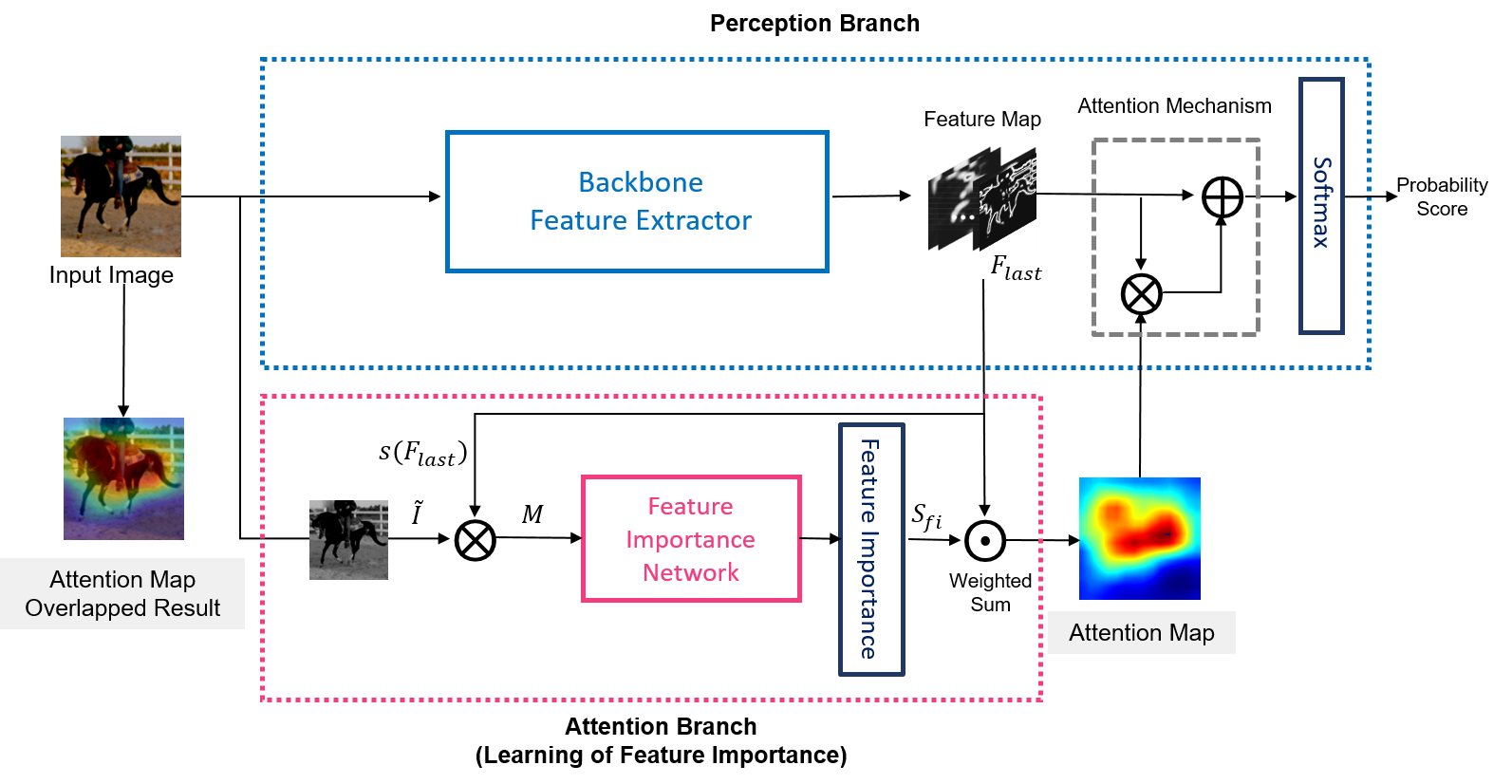}
\end{center}
  \caption{Overview of LFI-CAM Model.}
\label{fig:LFI-CAM}
\end{figure*}
Attention Branch Network (ABN)~\cite{fukui2019attention} was proposed not only to improve classification accuracy, but also to provide enhanced attention map for visual explanation simultaneously during inference time, by applying the attention mechanism. ABN is composed of the feature extractor, attention branch and perception branch.
To create an attention map, the attention branch generates $K \times h \times w$ feature map through multiple $1 \times 1$ convolution layers, and integrates the feature map into one channel by applying a single $1 \times 1$ convolution. Finally, the sigmoid function is applied to a $1 \times h \times w$ feature map for normalization. Here $K$ is the number of categories of the dataset and also the number of channels, while $h$ and $w$ are the height and width of the feature map.

We have observed that ABN outputs unreliable and inconsistent attention maps through several experiments. We trained several ABN models with various hyper-parameters on the Cat$\&$Dog dataset, and then compared CAMs of the same image from several models with similar accuracy.  As shown in Fig.~\ref{fig:Stability_Test}, CAM results for the exactly same test images are unreliable and inconsistent although the trained ABN models have similar accuracy.

We speculate that this phenomenon is caused by two reasons:

(1) The $K$-channel feature map generated from the attention branch becomes very shallow if the dataset's number of categories $K$ is small. The shallow feature map contributes to degrading the attention map quality and making the attention map inconsistent.

(2) The attention branch of ABN aggregates the $K$-channel feature map to a one-channel feature map without considering the channel-wise feature importance. Although the attention map is trained by the attention mechanism, it is possible to generate various types of attention maps depending on the changes in the initial weight parameters or hyper-parameters due to high degree of freedom.

\subsection{Score-CAM}

Score-CAM~\cite{wang2020score} is based on CAM with a simple but efficient importance representation for each feature map. Unlike previous gradient-based visual explanation approaches such as Grad-CAM~\cite{selvaraju2017grad} and Grad-CAM++~\cite{chattopadhay2018grad}, Score-CAM gets rid of the dependency on gradients by obtaining the weight of each feature map through its forward passing score on the target class. Ultimately, the final attention map is obtained as a weighted sum of feature maps.
In order to obtain the class-discriminative attention map of Score-CAM, each feature map is first up-sampled to the original input size and normalized to range [0, 1]. To project highlighted areas in the feature map to the original input space, a masked image $M^k$ is obtained by multiplying the normalized feature map $A^k$ with the original input $I$.
\begin{equation}
\begin{split} 
M^k=A^k\otimes I
\end{split} 
\label{eq:M_k}
\end{equation}
where $\otimes$ denotes element-wise multiplication and $k$ denotes the $k$-th channel of the last convolution layer. For each masked image $M^k$, the output score $S_k$ is obtained by the Softmax operation after forward computing $F(M^k)$.
\begin{equation}
\begin{split} 
S_k=Softmax(F(M^k))
\end{split} 
\label{eq:S_k}
\end{equation}
The score $S_k^c$ on target class $c$ represents the importance of the $k$-th feature map for target class $c$ which is $w_k^c$.
\begin{equation}
\begin{split} 
w_k^c=S_k^c
\end{split} 
\label{eq:w_k}
\end{equation}
The final class activation map is obtained by a linear weighted combination of all feature maps.
\begin{equation}
\begin{split} 
L_{Score-CAM}^c=ReLU(\sum_k w_k^c A^k)
\end{split} 
\label{eq:{Score-CAM}}
\end{equation}
Although Score-CAM achieves better visual performance with less noise and better stability than the gradient-based approaches, multiple forward computing makes the generation of visual explanation very slow.

\section{Proposed Method}

In this section, we introduce the proposed LFI-CAM which is trainable for image classification and visual explanation in an end-to-end manner. The LFI-CAM network is composed of the attention branch and perception branch, as shown in Fig.~\ref{fig:LFI-CAM}. The perception branch extracts feature maps from the input image by passing the input through multiple convolutional layers and predicts a class through the attention mechanism of the feature map from the convolutional layer and attention map. Meanwhile, the product between the feature maps and down-sampled grayscaled input is fed into the attention branch, also denoted as the Feature Importance Network (FIN). The FIN extracts the feature importance of each feature map, and then the weighted sum between the feature maps and extracted feature importance are calculated, generating the attention map. In this process, the attention branch not only generates an attention map for visual explanation by learning feature importance but also leverages the attention map to improve the classification performance through the attention mechanism.

\subsection{Motivation}
We propose a novel architecture, LFI-CAM, which is inspired by ABN~\cite{fukui2019attention} and Score-CAM~\cite{wang2020score}. By leveraging the attention mechanism of ABN, LFI-CAM improves the classification performance. However, the attention branch of ABN often generates unreliable and inconsistent visual explanation as elucidated above. To solve this issue, LFI-CAM's attention branch treats the feature maps as masks and obtains feature importance scores for each feature map to generate the attention map in a similar manner to Score-CAM. Since LFI-CAM's Feature Importance Network in the attention branch learns the feature importance for each feature map during training unlike Score-CAM, our attention map is generated much faster than that of Score-CAM during forward propagation.

\subsection{Feature Importance Network (Attention Branch)}

In contrast to the previous method~\cite{fukui2019attention}, which directly learns the class activation map in the attention branch, we replaced the ABN model's attention branch with a new network architecture, ``Feature Importance Network (FIN)". FIN helps the LFI-CAM model learn the feature importance to generate better class activation map. The class activation map is generated by the weighted sum of the feature maps from the last convolutional layer and the learned feature importance vector.

To learn the feature importance, FIN follows a similar approach to Score-CAM~\cite{wang2020score}. However, unlike Score-CAM, we convert the original input into a gray image, which is downsampled to the size of the feature map. In addition, instead of conducting several forward computations, a concatenated masked image is fed as an input into the FIN. The feature importance for each feature map is outputted from the FIN.

In order to obtain the class activation map of LFI-CAM, the original input $I \in \mathbb{R}^{3\times w \times h}$  is first  converted from RGB color space to a single gray scale space and down-sampled to a feature map with a size of $\mathbb{R}^{1 \times m \times n}$. An example would be a conversion from  $I \in \mathbb{R}^{3\times 224 \times 224}$ to  $\tilde{I} \in \mathbb{R}^{1\times 14 \times 14}$ in Resnet18~\cite{he2016deep} architecture.
\begin{equation}
\begin{split} 
\tilde{I}=Down(rgb2gray(I))
\end{split} 
\label{eq:{tildeI}}
\end{equation}
 
 Each feature map of the last convolutional layer, $F^k_{last} \in \mathbb{R}^{1 \times m \times n}$ is normalized, where $k$ denotes the channel index of the last convolutional layer. A masked image $M_k \in \mathbb{R}^{1 \times m \times n}$ is obtained by multiplying the down-sampled gray input image $\tilde{I}$ with the normalized feature map.
\begin{equation}
\begin{split} 
M_k=\tilde{I} \otimes s(F^k_{last})
\end{split} 
\label{eq:M_k-LFI-CAM}
\end{equation}
 where $s(\cdot)$ is a normalization function that  maps each element in every feature map to range [0,1]. We generate a set of masked images $\{M_1, M_2, ..., M_N\}$ and concatenate them all, where $N$ is the number of channels of the last convolutional layer of the model. Finally, we feed the concatenated masked image $M$ into the FIN model $FIN(x)$ to conduct a forward propagation $FIN(M)$.
\begin{equation}
\begin{split} 
S_{fi} = FIN(M)
\end{split} 
\label{eq:S_{fi}}
\end{equation}
where $S_{fi} \in \mathbb{R}^{N}$ is the feature importance score vector. We take the $k$-th score $S_{fi}^k$ as weight to represent the feature importance of the $k$-th feature map.
\begin{equation}
\begin{split} 
w_k = S_{fi}^k
\end{split} 
\label{eq:w_k}
\end{equation}

The class activation map of the LFI-CAM is obtained by a weighted combination of all feature maps.
\begin{equation}
\begin{split} 
L_{LFI-CAM}=ReLU(\sum_{k=1}^{N} w_k F^k_{last})
\end{split} 
\label{eq:L_{LFI-CAM}}
\end{equation}

Similar to previous works~\cite{selvaraju2017grad, wang2020score, chattopadhay2018grad}, ReLU is applied to the linear combination of feature maps to remove  features with negative influence.

\begin{figure*}
\begin{center}
\includegraphics[width=1\linewidth, height=0.85\linewidth]{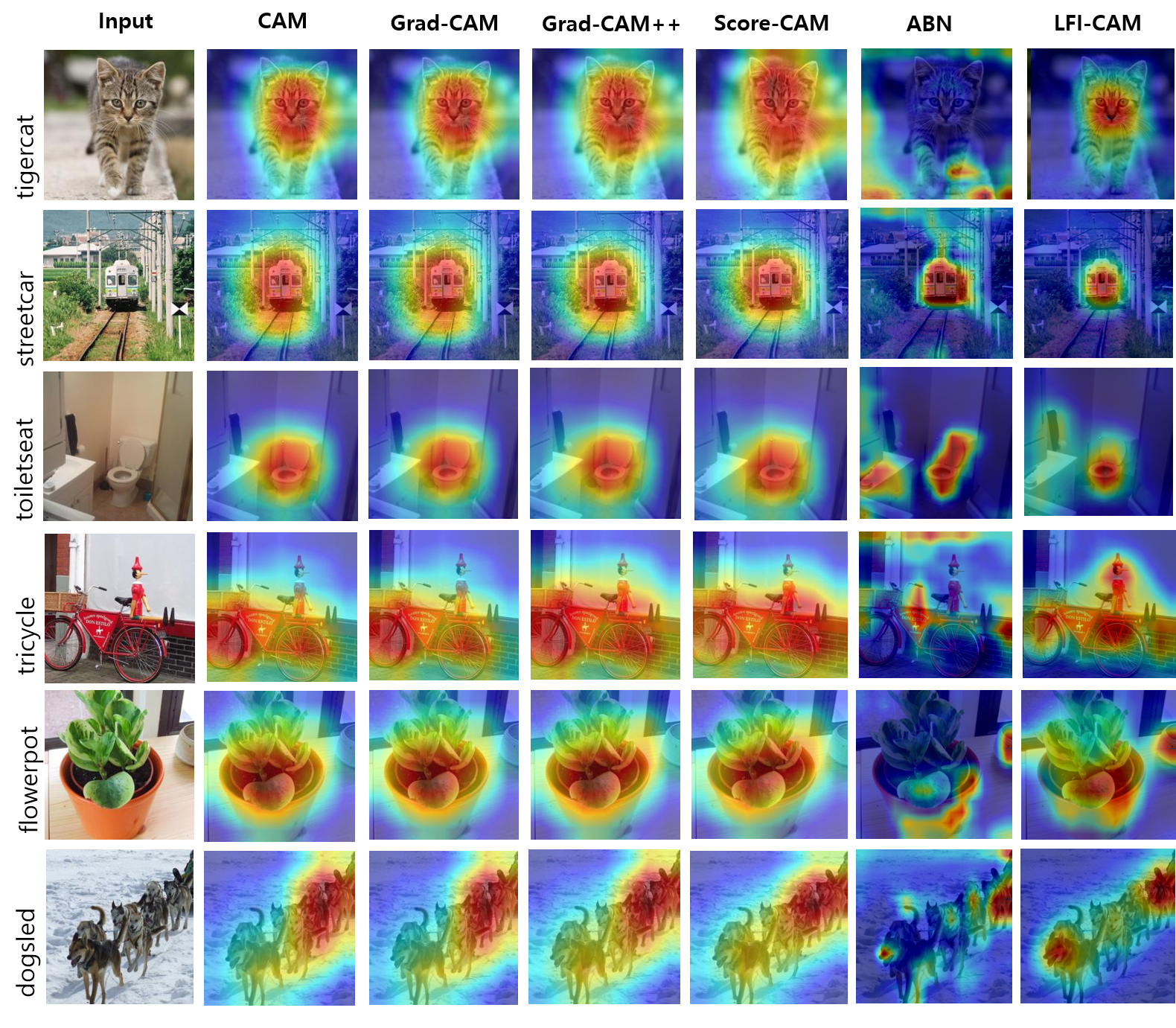}
\end{center}
\vspace{-5mm}
  \caption{Visual explanation results of various methods. More results are provided in the supplementary material.}
\label{fig:SingleObject}
\end{figure*}

\subsection{Perception Branch}

The perception branch takes the original input image as an input and outputs the final probability of each class. The attention map generated from the attention branch (FIN) is overlapped onto the feature maps from the intermediate convolutional layer by the attention mechanism. Unlike ABN, the LFI-CAM attention map is always generated using feature maps from the last convolutional layer, instead of the feature extractor. However, the attention mechanism can be applied to the feature map from any convolutional layer. We use the following attention mechanism formula from ABN~\cite{fukui2019attention}.
\begin{equation}
\begin{split} 
\acute{F_l^k} = (1+ L_{LFI-CAM})\otimes F_l^k
\end{split} 
\label{eq:F_l^k}
\end{equation}
where $F_l^k$ is the feature map at the $l$-th convolutional layer and $\acute{F_l^k}$ is the output of the attention mechanism. Note that $k$ is the index of the channel and that $L_{LFI-CAM}$ is normalized to range [0,1] before being used in the attention mechanism. The attention mechanism helps the attention map improve the classification performance by highlighting the feature map at the location with a higher value of attention map while preventing the lower value region of the attention map from degrading to zero.

\vspace{-2mm}
\subsection{Training}

LFI-CAM is trained in an end-to-end manner using training loss calculated as the combination of the Softmax function and cross-entropy at the perception branch in image classification task. The FIN is optimized by the attention mechanism of the perception branch to improve the classification accuracy without any additional loss function.

\begin{figure*}
\begin{center}
\includegraphics[width=1\linewidth, height=0.45\linewidth]{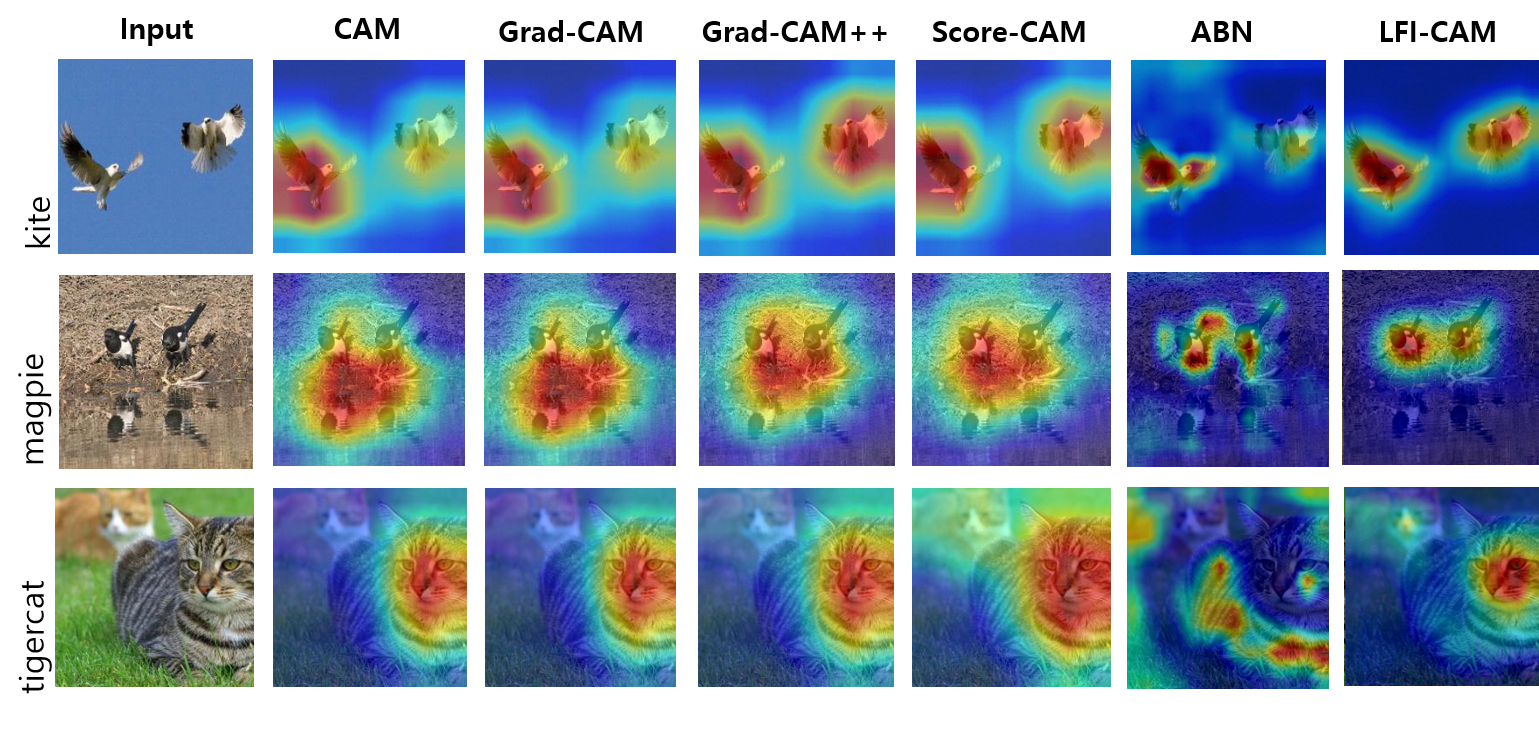}
\end{center}
  \caption{Visual explanation results of various methods for multi-target. More results are provided in the supplementary material.}
\label{fig:Multi-Object}
\end{figure*}
\section{Experiments}

In this section, we evaluate LFI-CAM's classification performance and explain the effectiveness. First, we describe the experimental settings on image classification in Sec 4.1. Second, we qualitatively evaluate our approach via visualization on ImageNet in Sec 4.2. In Sec 4.3, we quantitatively evaluate LFI-CAM's image classification performance by comparing it with various baseline models. Finally, we measure the stability of LFI-CAM's visual explanation and compare it with the stability of ABN's visual explanation.

\subsection{Experimental Settings on Image Classification}
{\bf Datasets:} We evaluate LFI-CAM using 5 different public datasets- CIFAR10, CIFAR100, STL10, Cat$\&$Dog (Kaggle Cats and Dogs), and ImageNet~\cite{deng2009imagenet}. The Cat$\&$Dog dataset has 2 classes, CIFAR10 and STL10 dataset have 10 classes each, CIFAR100 dataset has 100 classes, and the ImageNet dataset has 1,000 classes. Training and testing dataset sizes are as follows: CIFAR10 and CIFAR100 has 50,000 training images and 10,000 testing images, and ImageNet consists of 1,281,167 training images and 50,000 testing images. STL10 consists of 5,000 training images and 8,000 testing images. The Cat$\&$Dog dataset has 8,007 training images and 2,025 testing images. The input image size of CIFAR10 and CIFAR100 datasets is 32 x 32 pixels, for STL10 it is 96 x 96, for Cat$\&$Dog dataset and ImageNet it is 224 x 224 pixels.

{\bf Base Models:} In this experiment, CIFAR10 and CIFAR100 are evaluated via the CIFAR ResNet backbone (ResNet {20, 32, 44, 56, 110}). The STL10, Cat$\&$Dog, and ImageNet dataset is evaluated via the ImageNet ResNet backbone (ResNet {18, 34, 50, 101, 152}). The CIFAR ResNet backbone is more lightweight than the ImageNet ResNet backbone, with fewer layers and parameters, which is suitable for the relatively smaller sized input image. The CIFAR ResNet backbone uses standard data augmentation of zero-padding images with 4 pixels on each side and then randomly cropping to produce 32 x 32 pixels images. Subsequently, horizontal flip is applied at random. For ImageNet ResNet backbone, the training images are randomly resized and cropped to 224 x 224 pixels and then horizontally mirrored at random. The validation images are resized to 256 x 256 and then center cropped to produce 224 x 224 sized images. All LFI-CAM models are composed of the perception branch (backbone) and attention branch (FIN), where the FIN is constructed with multiple convolutional layers. Further details on the LFI-CAM architecture can be found in the supplementary material.

{\bf Optimizer and Hyper-parameters:} We use the most standard optimizer which is stochastic gradient descent (SGD) with momentum. We set the total epoch hyper-parameter as follows: CIFAR10, CIFAR100, STL10, and Cat$\&$Dog are 300 epochs, and ImageNet is 90 epochs. The learning rate is initialized with 0.1, and later on divided by 10 at 50 $\%$ and 75 $\%$ of the total number of training epochs. We used training batch size of 128 for CIFAR ResNet backbone and 256 for ImageNet ResNet backbone.

\subsection{Visual Explanation Evaluation}

We qualitatively compare the visual explanation generated by 5 state-of-the-art methods, namely CAM~\cite{zhou2016learning}, Grad-CAM~\cite{selvaraju2017grad}, Grad-CAM++~\cite{chattopadhay2018grad}, ScoreCAM~\cite{wang2020score} and ABN~\cite{fukui2019attention}. While CAM, Grad-CAM, Grad-CAM++ and Score-CAM can generate class activation map for each target class, ABN and LFI-CAM always generate a single class activation map for the class with the highest prediction probability. To make the comparison as fair as possible, we used the predicted output class of LFI-CAM as the target class for CAM, Grad-CAM, Grad-CAM++, and Score-CAM and selected examples with the same prediction result for both LFI-CAM and ABN. For CAM, Grad-CAM, Grad-CAM++, and ScoreCAM, we used ResNet18 model pretrained on ImageNet. For ABN and LFI-CAM, we used ResNet18 backbone trained on ImageNet.  

\subsubsection{Class Discriminative Visualization}
  
  As can be seen in Fig.~\ref{fig:SingleObject}, our method shows high-quality results beyond equivalence compared to CAM-variant methods, especially demonstrating less noisy and more focused results on target object area. Our approach can also generate much more reliable visual explanation compared to ABN. More examples are provided in the supplementary material.

LFI-CAM shows better performance on locating multiple target objects than previous works as shown in Fig.~\ref{fig:Multi-Object}. ABN often shows noisy and unreliable results as most attention maps are not generated properly or are generated in unrelated areas. Compared to other CAM-variant methods, our approach yields more focused and less noisy results as shown in the single object experiment.

\begin{figure}
\begin{center}
\includegraphics[width=1\linewidth, height=0.6\linewidth]{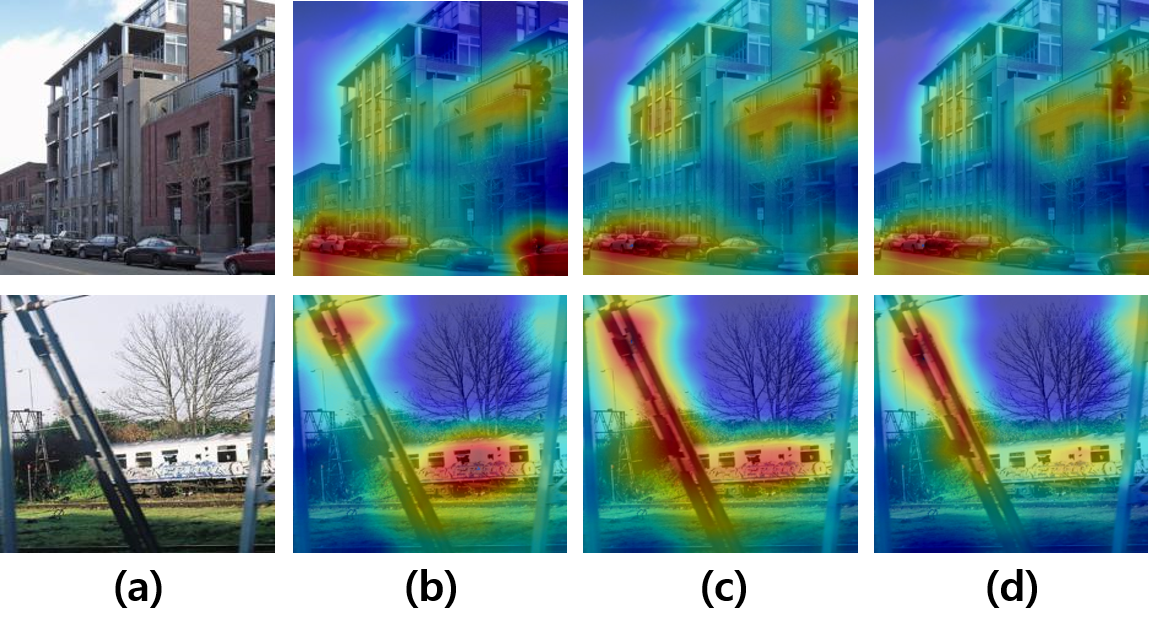}
\end{center}
  \caption{Visualization of the Feature Importance Network Effectiveness.  (a) Input image, (b) Pixel-wise mean feature map from the last convolutional layer of the LFI-CAM model trained without FIN, (c) Pixel-wise mean feature map from the last convolutional layer of the LFI-CAM model trained with FIN, (d) CAM generated from the LFI-CAM model. }
\label{fig:FIN_Effect}
\end{figure}

\begin{figure}
\begin{center}
\includegraphics[width=1\linewidth, height=1.2\linewidth]{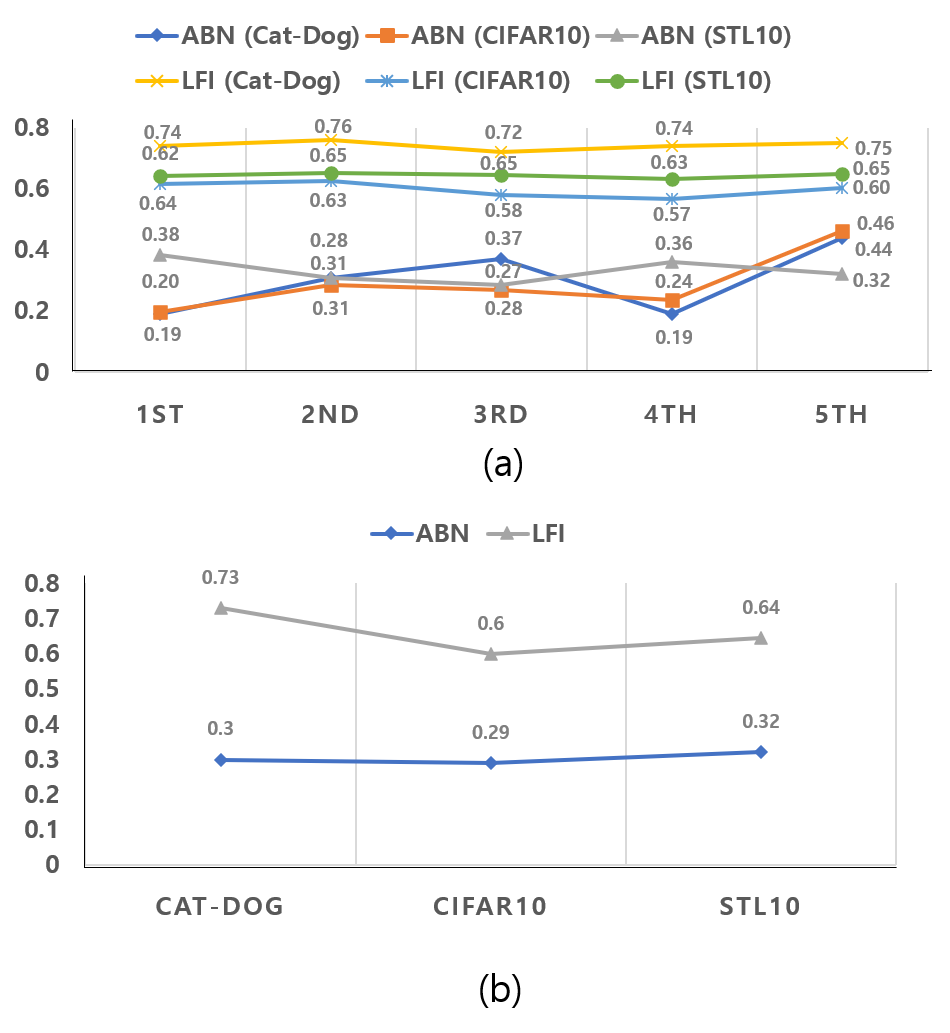}
\end{center}
\vspace{-5mm}
  \caption{Stability evaluation of visual explanation. (a) IoU between models per dataset,                            (b) Average IoU per dataset.}
\label{fig:IOU_Test}
\end{figure}

\begin{figure}
\begin{center}
\includegraphics[width=1\linewidth, height=1\linewidth]{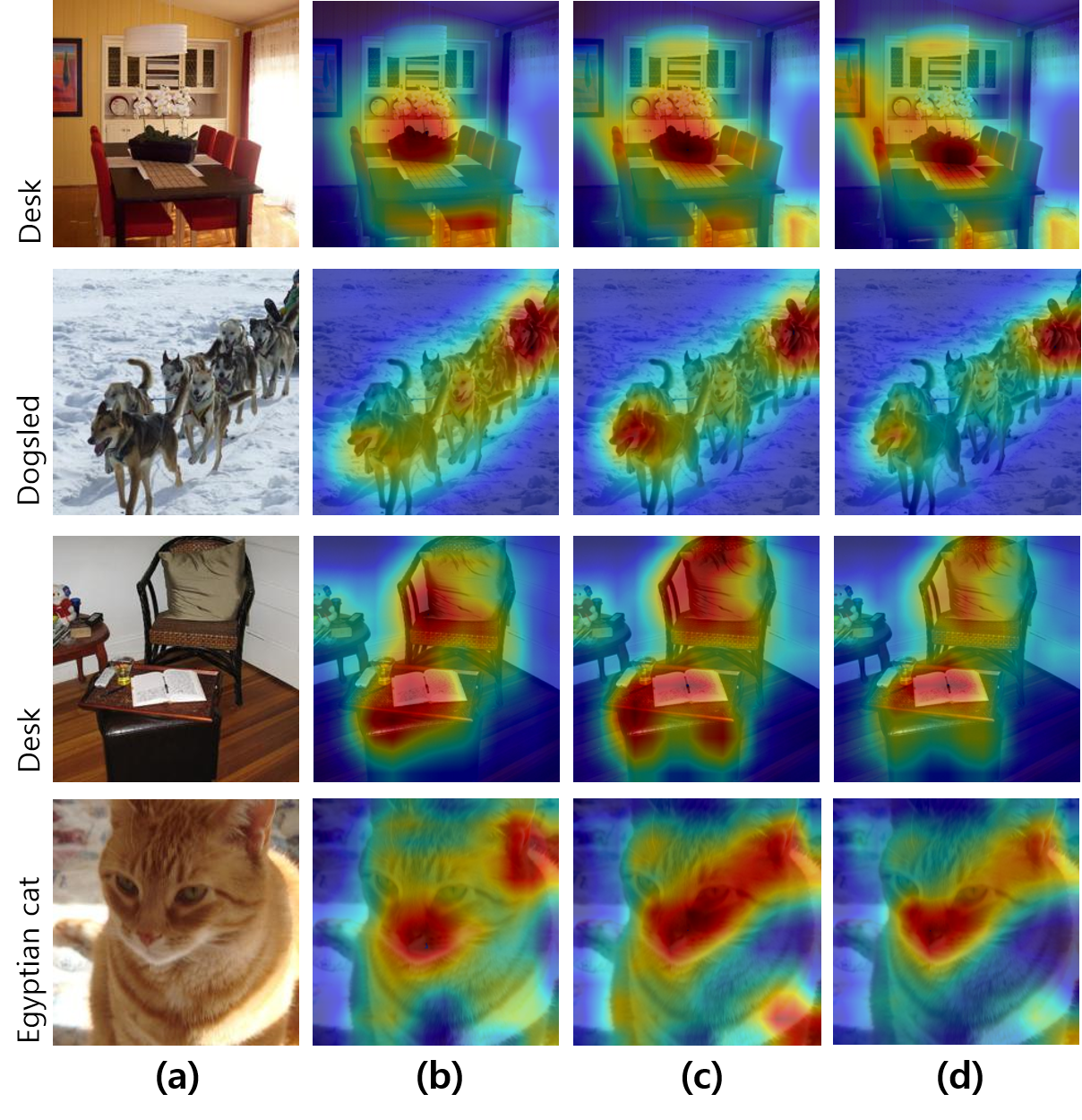}

\end{center}
  \caption{Visualization of the Feature Importance Network Effectiveness.  (a) Input image, (b) Pixel-wise mean feature map from the last convolutional layer of the LFI-CAM model trained without FIN, (c) Pixel-wise mean feature map from the last convolutional layer of the LFI-CAM model trained with FIN, (d) CAM generated from the LFI-CAM model. }
\label{fig:FIN_Effect_supple}
\end{figure}

\begin{table}[h]
\caption{Comparison of top-1 errors on CIFAR10 and CIFAR100.}
\label{table1}
\centering
\begin{center}
\begin{tabular}{|c||c|c|} \hline
{Model} &  CIFAR10 & CIFAR100 \\
\hline\hline
ResNet110 & 6.43 & 24.14   \\ 
ResNeXt~\cite{xie2017aggregated}  & 3.84 & 18.32  \\ \hline
ResNet110+ABN  & $\textbf{4.91}_{(-1.52)}$ & $\textbf{22.82}_{(-1.32)}$  \\
ResNeXt+ABN  & $\textbf{3.8}_{(-0.04)}$ & $\textbf{17.7}_{(-0.62)}$  \\ \hline 
ResNet110+LFI-CAM  & $5.73_{(-0.7)}$  & $23.33_{(-0.81)}$  \\
ResNeXt+LFI-CAM & $4.27_{(+0.43)}$ & $18.23_{(-0.09)}$ \\ \hline 

\end{tabular}
\end{center}
\end{table}

\begin{table}[h]
\caption{Comparison of top-1 errors on STL10 and Cat$\&$Dog.}
\label{table2}
\centering
\begin{center}
\begin{tabular}{|c||c|c|} \hline
{Model} &  STL10 & Cat$\&$Dog \\
\hline\hline
ResNet18 +ABN & 18.75 & 3.07 \\
ResNet18+LFI-CAM & $\textbf{18.16}_{(-0.59)}$ & $\textbf{2.72}_{(-0.35)}$ \\ \hline

\end{tabular}
\end{center}
\end{table}

\begin{table}[h]
\caption{Comparison of top-1 errors on ImageNet dataset.}
\label{table3}
\centering
\begin{center}
\begin{tabular}{|c||c|c|} \hline
{Model} & {Model Size} & ImageNet \\
\hline\hline

ResNet18 & 11.17M & 30.24 \\
ResNet34 & 21.28M & 26.69 \\
ResNet50 & 23.25M & 23.87 \\
ResNet101 & 42.51M & 22.63 \\
ResNet152 & 58.16M &  21.69 \\ \hline

ResNet18+ABN & $21.61M_{(+10.44)}$ & $28.98_{(-1.26)}$ \\ 
ResNet34+ABN & $36.44M_{(+15.16)}$ & $25.78_{(-0.91)}$ \\ 
ResNet50+ABN & $43.58M_{(+20.33)}$ & $23.1_{(-0.77)}$\\ 
ResNet101+ABN & $62.58M_{(+20.07)}$ & $\textbf{21.8}_{(-0.83)}$ \\ 
ResNet152+ABN & $78.22M_{(+20.06)}$ & $\textbf{21.4}_{(-0.29)}$ \\ \hline

ResNet18+LFI-CAM & $\textbf{17.47M}_{(+6.3)}$ & $\textbf{27.75}_{(-2.49)}$ \\
ResNet34+LFI-CAM & $\textbf{29.94M}_{(+8.66)}$ & $\textbf{25.68}_{(-1.01)}$ \\
ResNet50+LFI-CAM & $\textbf{43.05M}_{(+19.8)}$ & $\textbf{22.71}_{(-1.16)}$\\
ResNet101+LFI-CAM & $\textbf{62.04M}_{(+19.53)}$ & $21.84_{(-0.79)}$ \\
ResNet152+LFI-CAM & $\textbf{77.68M}_{(+19.52)}$ & $21.95_{(+0.26)}$ \\ \hline

\end{tabular}
\end{center}
\end{table}

\begin{figure*}
\begin{center}
\includegraphics[width=1\linewidth, height=0.25\linewidth]{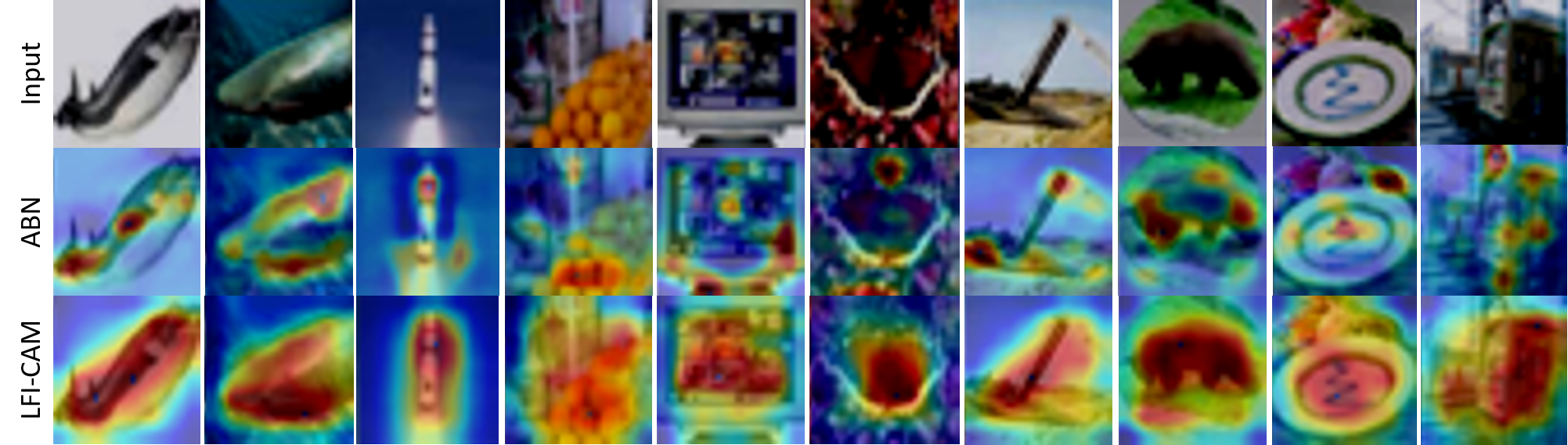}
\end{center}
  \caption{Visual Explanation Results of ABN and LFI-CAM for CIFAR100}
\label{fig:Comparison_ABN_LFICAM_CIFAR100}
\end{figure*}

\begin{figure*}
\begin{center}
\includegraphics[width=1\linewidth, height=0.25\linewidth]{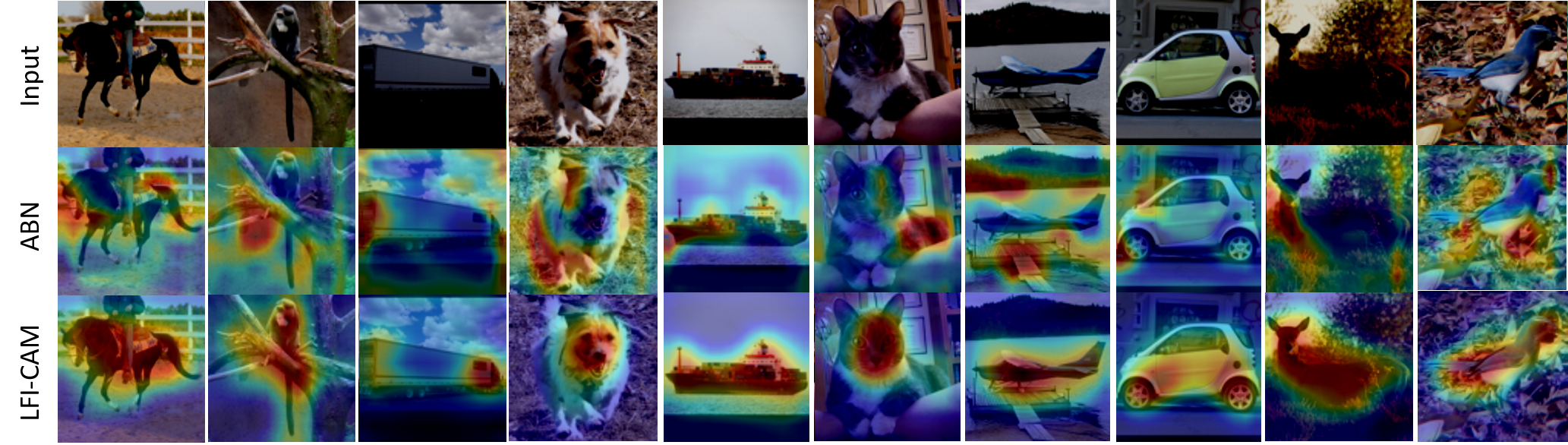}
\end{center}
  \caption{Visual Explanation Results of ABN and LFI-CAM for STL10}
\label{fig:Comparison_ABN_LFICAM_STL10}
\end{figure*}

\begin{figure*}
\begin{center}
\includegraphics[width=1\linewidth, height=0.25\linewidth]{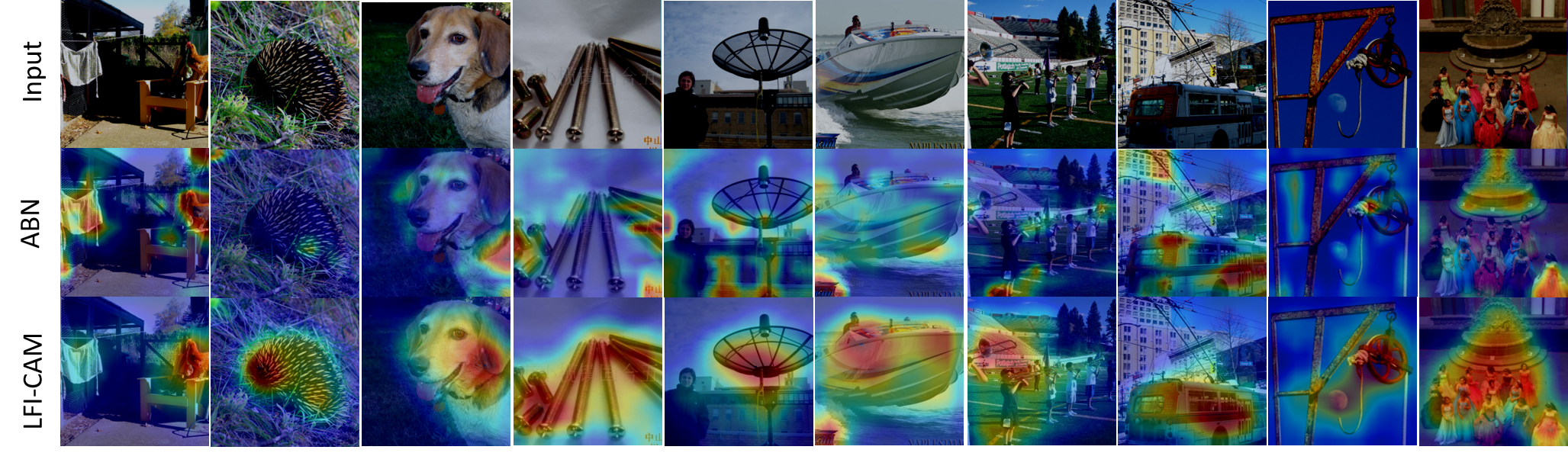}
\end{center}
  \caption{Visual Explanation Results of ABN and LFI-CAM for ImageNet}
\label{fig:Comparison_ABN_LFICAM_ImageNet}
\end{figure*}

\subsubsection{Effectiveness of Feature Importance Network}

To evaluate the effectiveness of the proposed Feature Importance Network, we visualize the pixel-wise mean feature map from the last convolutional layer of the LFI-CAM model trained without and with the FIN. Then we compare them against the CAM generated from LFI-CAM model trained with FIN. Although we initially expected that the feature importance learned by the FIN plays an important role in generating reliable CAM, an interesting discovery is that the backbone network, FIN, and attention mechanism interact with each other during training. Therefore, we confirmed that LFI-CAM model is optimized not only by learning the feature importance but also by enhancing the backbone feature representation to focus more on important features to make decision for the input image. As shown in Fig.~\ref{fig:FIN_Effect}, after the FIN's feature importance is incorporated, our $L_{LFI-CAM}$ successfully focuses on the most distinguishable region of the target object. For example, as shown in the second row, the steel structure is highlighted prominently after applying the FIN because the LFI-CAM model classifies the input image as `pole'.

\subsection{Accuracy on Image Classification}

\textbf{Accuracy on CIFAR10 and CIFAR100:}
Table~\ref{table1} shows the top-1 errors on CIFAR10/100. We evaluate these top-1 errors using ResNet110, ResNeXt, ABN and LFI-CAM. The numbers in brackets denote the difference in top-1 errors from the conventional models, ResNet110 and ResNeXt. Although LFI-CAM outperforms the conventional models, ABN's top-1 errors tend to be slightly smaller than LFI-CAM's top-1 errors. However, we confirmed that LFI-CAM's CAM is much more reliable than ABN's CAM. The attention maps of ABN and LFI-CAM are provided in the supplementary material. 

\textbf{Accuracy on STL10 and Cat$\&$Dog:}
We evaluate the image classification accuracy on STL10 and Cat$\&$Dog as shown in Table~\ref{table2} with the same method used for CIFAR10/100. We evaluate the top-1 errors for ResNet18 with ABN and LFI-CAM. On STL10, ResNet with LFI-CAM decreases the top-1 errors by 0.59 compared to ResNet18 with ABN. On  Cat$\&$Dog,  LFI-CAM also decreases the top-1 errors by 0.35 compared to ResNet18 with ABN.

\textbf{Accuracy on ImageNet:}
We evaluate the image classification accuracy on ImageNet as shown in Table~\ref{table3}. We tested ResNet{18, 34, 50, 101, 152} models with ABN and LFI-CAM. The table shows that LFI-CAM is on par with ABN in terms of classification accuracy for each backbone model, even with less model parameters. Furthermore, LFI-CAM generates much more reliable CAM than ABN as shown in Sec 4.2.1.

\subsection{Stability Evaluation of Visual Explanation}

The stability of visual explanation is an important measure of CAM-related algorithm's performance and real world applicability. Researchers have observed instability of visual explanation from several previous works  ~\cite{selvaraju2017grad, wang2020score}, and one recent work, the Attention Branch Network~\cite{fukui2019attention}, shows significant instability of visual explanation for datasets with fewer number of classes, such as Cat$\&$Dog, CIFAR10, STL10. Hence, we evaluated the stability of visual explanation performance of LFI-CAM and other relevant models with those datasets. As shown in Fig.~\ref{fig:Stability_Test}, we observed that LFI-CAM shows stable visual explanatory performance unlike previous works such as ABN. As a measure of stability, we used the IoU (Intersection of Union) between visual explanations on all the test data generated by six models trained separately on the same dataset. First, we select one model with the highest classification accuracy from the 6 models as a baseline. Then, we compare the IoU between the visual explanation generated from the other 5 models with the baseline model. Since the image area where the model gives more attention will have higher temperature, we used visual explanations with reasonably high temperature ($\geq$127, value range [0, 256]) for IoU calculation. As seen in Fig.~\ref{fig:IOU_Test}, when comparing areas with high attention, LFI-CAM shows that 60$\%$ or more overlap on average, but for ABN, 30$\%$ overlap. The backbone network used for stability evaluation was ResNet18 for Cat$\&$Dog and STL10 dataset, and ResNet110 for Cifar10 dataset. 

\section{Conclusion and Future Work}
In this paper, we proposed the LFI-CAM model, which is trainable for image classification and produces better visual explanation in an end-to-end manner. We replaced the ABN model's attention branch with a new network architecture that we call ``Feature Importance Network (FIN)" which helps our model focus on learning the feature importance to generate a more stable and reliable attention map. In other words, LFI-CAM's attention map is generated by the weighted sum of the feature map from the last convolutional layer and the learned feature importance, while ABN learns the attention map itself without taking the feature importance into consideration. Throughout the paper, we evaluated the classification performance and visual explanation quality of LFI-CAM, and we concluded that LFI-CAM is on par with ABN in terms of classification accuracy and outmatches ABN in terms of attention map quality. Future work is planned to apply LFI-CAM's Feature Importance Network to other tasks such as object detection, semantic segmentation and multi-task learning. 

\section{Network Architecture}
The LFI-CAM network is composed of the attention branch and perception branch and both branches are connected by the attention mechanism. The conventional baseline models such as ResNet, DenseNet, ResNeXt and SENet or the customized classifier can be used as the perception branch. The Feature Importance Network(FIN) contains multiple convolution layers for extracting feature importance of the feature map. The architecture details of the FIN are shown in Table~\ref{table1}. There are some notations: $h$ and $w$: height and width of the input image, N: the number of output channels, K:kernel size, S:stride size, P:padding size, BN:batch normalization.

\section{Additional Experimental Results}

\subsection{Visual Explanation Results}
In addition to the results presented in the paper, we show supplement visual explanation results of ABN and LFI-CAM for the CIFAR100, STL10 and ImageNet datasets in Figs.~\ref{fig:Comparison_ABN_LFICAM_CIFAR100}, Figs.~\ref{fig:Comparison_ABN_LFICAM_STL10} and Figs.~\ref{fig:Comparison_ABN_LFICAM_ImageNet}

\subsection{Visualization of the Feature Importance Network Effectiveness}
To evaluate the effectiveness of the proposed Feature Importance Network, we visualize the pixel-wise mean feature map from the last convolutional layer of the LFI-CAM model trained without and with the FIN. Then we compare them against the CAM generated from LFI-CAM model trained with FIN. Fig.~\ref{fig:FIN_Effect_supple} shows the additional visual explanation results for FIN effectiveness. After the FIN's feature importance is incorporated, our $L_{LFI-CAM}$ successfully focuses on the most distinguishable region of the target object. For example, as shown in the first and third row, the attention focuses more on the desk area after applying the FIN because the LFI-CAM model classifies the input image as `desk'.

\subsection{Stability Evaluation on Visual Explanation for ABN and LFI-CAM}

\begin{figure*}
\begin{center}
\includegraphics[width=0.8\linewidth, height=1.0\linewidth]{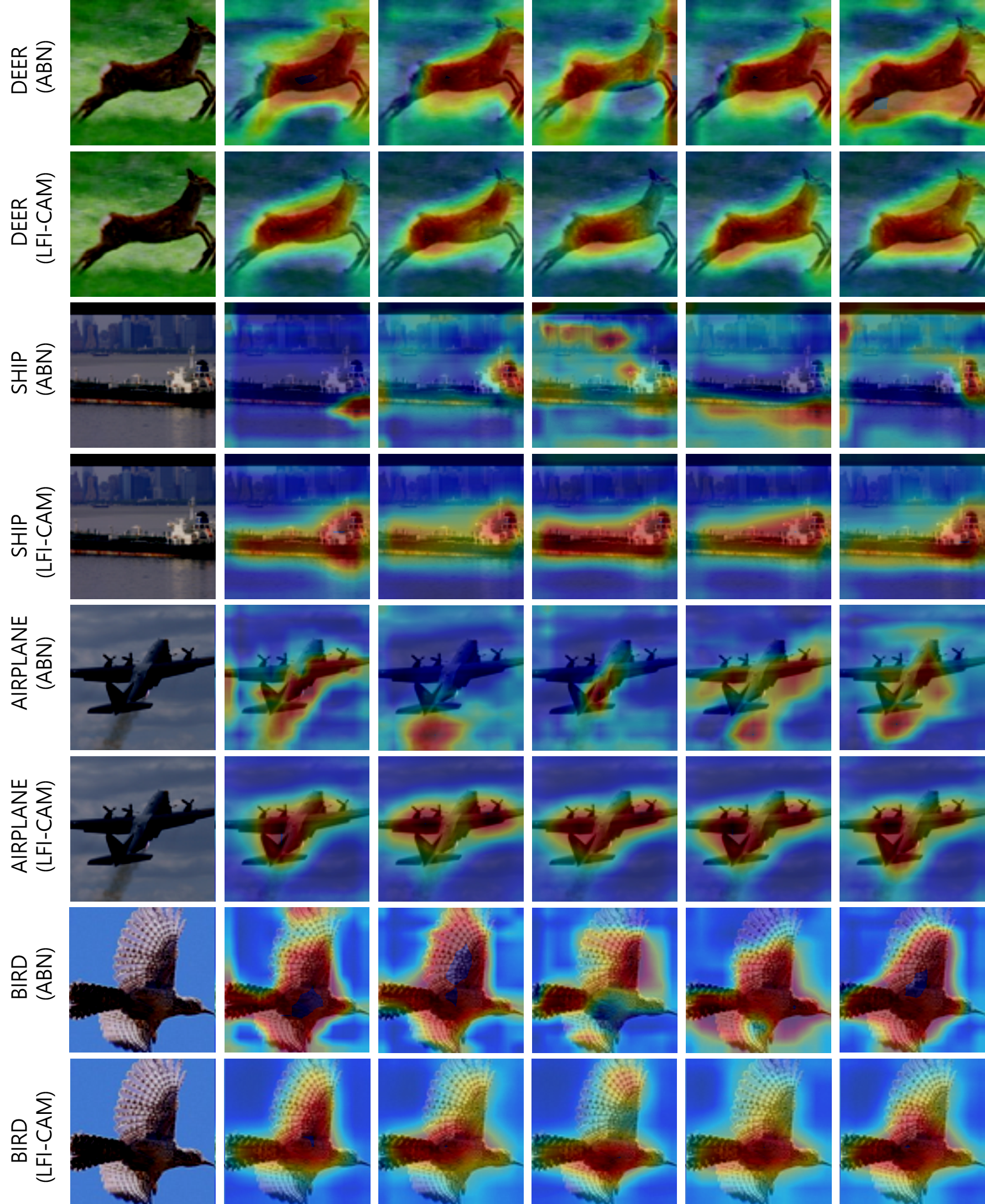}
\end{center}
  \caption{Examples of stability test on visual explanation. Each row displays CAM results of ABN or LFI-CAM models that were trained with various (5) hyper-parameters on the STL10 dataset.}
\label{fig:STL10_ABN_vs_LFI}
\end{figure*}

\begin{figure*}
\begin{center}
\includegraphics[width=0.8\linewidth, height=1.0\linewidth]{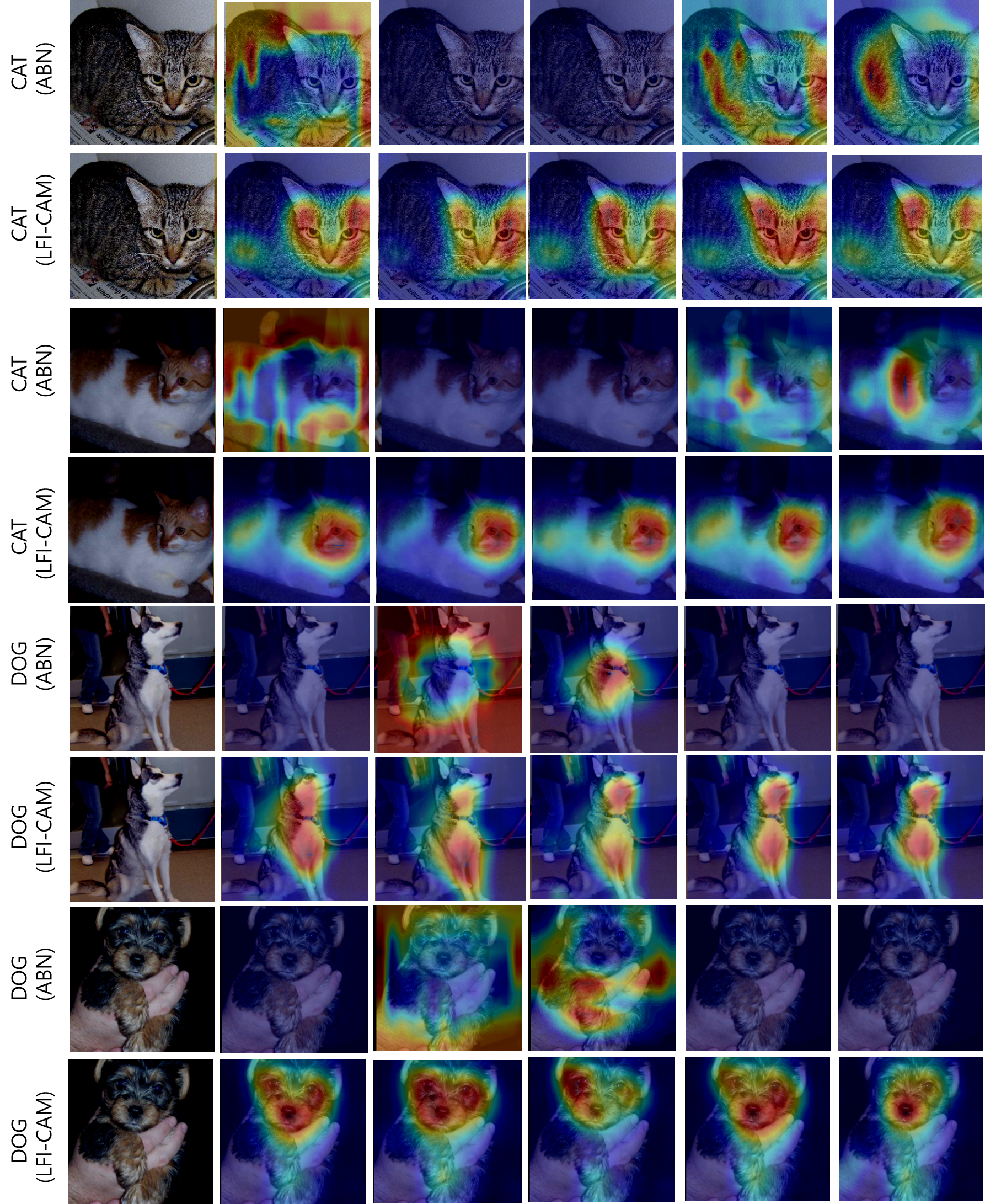}
\end{center}
  \caption{Examples of stability test on visual explanation. Each row displays CAM results of ABN or LFI-CAM models that were trained with various (5) hyper-parameters on the Cat$\&$Dog dataset.}
\label{fig:CATDOG_ABN_vs_LFI}
\end{figure*}

We have observed that ABN outputs unreliable and inconsistent attention maps through several experiments. We trained several ABN models with various hyper-parameters on the Cat$\&$Dog and STL10 dataset, and then compared CAMs of the same image from several models with similar accuracy.  
Fig.~\ref{fig:STL10_ABN_vs_LFI} and Fig.~\ref{fig:CATDOG_ABN_vs_LFI} show the additional stability test results on visual explanation for the STL10 and Cat$\&$Dog dataset. CAM results for the exactly same test images are unreliable and inconsistent although the trained ABN models have similar accuracy. On the other hand, the results of LFI-CAM can be confirmed to be reliable and stable.


\begin{table*}[h]
\caption{Architecture of Feature Importance Network (FIN).}
\label{table1}
\centering
\begin{center}
\begin{tabular}{|c||c|c|} \hline
Part & Input $\rightarrow$ Output Shape & Layer Information \\
\hline\hline
CONV Layer & ($\frac{h}{16}$, $\frac{w}{16}$,2048) $\rightarrow$ ($\frac{h}{16}$, $\frac{w}{16}$, 2048) & CONV-(N2048, K3, S0, P1), BN, ReLU \\ \cline{2-3}
& ($\frac{h}{16}$, $\frac{w}{16}$,2048) $\rightarrow$ ($\frac{h}{16}$, $\frac{w}{16}$, 2048) & CONV-(N2048, K3, S0, P1), BN, ReLU \\ \cline{2-3}
& ($\frac{h}{16}$, $\frac{w}{16}$,2048) $\rightarrow$ ($\frac{h}{16}$, $\frac{w}{16}$, 2048) & CONV-(N2048, K3, S0, P1), BN, ReLU \\ \cline{2-3}
& ($\frac{h}{16}$, $\frac{w}{16}$,2048) $\rightarrow$ ($\frac{h}{16}$, $\frac{w}{16}$, 2048) & CONV-(N2048, K3, S0, P1), BN, ReLU \\ \hline

Output Layer & ($\frac{h}{16}$, $\frac{w}{16}$, 2048) $\rightarrow$ (2048) & Global Average Pooling $\&$ SoftMax \\ \hline
 
\end{tabular}
\end{center}
\end{table*}

{\small
\bibliographystyle{ieee_fullname}
\bibliography{egbib}
}

\end{document}